\documentclass[letterpaper, 10 pt, conference]{ieeeconf}
\IEEEoverridecommandlockouts 
\usepackage[nolist]{acronym}
\usepackage{bm}
\usepackage{booktabs}
\usepackage{color}
\usepackage{cite}
\usepackage{graphicx}
\usepackage{multirow}
\usepackage{tabularx}
\usepackage{siunitx}

\usepackage[normalem]{ulem}
\let\labelindent\relax
\usepackage{enumitem}
\usepackage{url}
\usepackage{hyperref} 

\newacro{hri}[HRI]{Human-Robot Interaction}
\newacro{lstm}[LSTM]{Long Short-Term Memory}

\newcommand{\diff}[1]{{\color{black}#1}}

\title{\LARGE \bf
A Service Robot in the Wild: Analysis of Users Intentions,\\Robot Behaviors, and Their Impact on the Interaction
}
\author{Simone Arreghini, Gabriele Abbate, Alessandro Giusti and Antonio Paolillo
\thanks{This work was supported by the European Union through the project SERMAS and by the Swiss State Secretariat for Education, Research and Innovation (SERI) under contract number 22.00247.}
\thanks{All the authors are with Dalle Molle Institute for Artificial Intelligence (IDSIA), USI-SUPSI, Lugano, Switzerland {\tt name.surname@idsia.ch}
\linebreak
~
\linebreak
\vspace{-0.75mm}
\noindent\fbox{\footnotesize\begin{minipage}{0.99\textwidth}\copyright 2024 IEEE.  Personal use of this material is permitted. Permission from IEEE must be obtained for all other uses, in any current or future media, including reprinting/republishing this material for advertising or promotional purposes, creating new collective works, for resale or redistribution to servers or lists, or reuse of any copyrighted component of this work in other works.
\end{minipage}}
\vspace{-1.5cm}%
}%
}

\begin{document}

\maketitle
\thispagestyle{empty}
\pagestyle{empty}

\begin{abstract}
We consider a service robot that offers chocolate treats to people passing in its proximity: it has the capability of predicting in advance a person's intention to interact, and to actuate an ``offering'' gesture, subtly extending the tray of chocolates towards a given target.
We run the system for more than 5 hours across 3 days and two different crowded public locations; the system implements three possible behaviors that are randomly toggled every few minutes: passive (e.g. never performing the offering gesture); or active, triggered by either a naive distance-based rule, or a smart approach that relies on various behavioral cues of the user. 
We collect a real-world dataset that includes information on 1777 users with several spontaneous human-robot interactions and study the influence of robot actions on people's behavior.
Our comprehensive analysis suggests that users are more prone to engage with the robot when it proactively starts the interaction.
We release the dataset and provide insights to make our work reproducible for the community.
Also, we report qualitative observations collected during the acquisition campaign and identify future challenges and research directions in the domain of social human-robot interaction.
\end{abstract}
\section{Introduction}\label{sec:intro}

Recent robotics applications foresee close interaction
with humans in environments such as schools~\cite{Benitti:cae:2012}, hospitals~\cite{Gonzalez:as:2021}, and hotels~\cite{Choi:jhmm:2020}.
%
In such contexts, robots are now required to perform gently and comfortably when dealing with human users, manifest emotional~\cite{sirithunge2019proactive} or social intelligence~\cite{Zaraki:icrm:2014,nocentini2019survey}, and be able to comply with people's needs and attitudes. 
For this reason, \emph{social} \ac{hri} is quickly spreading to develop meaningful and safe coexistence with people in everyday life scenarios. 

\begin{figure}[t]
\centering%
\frame{\includegraphics[trim={2.5cm 0.5cm 2.5cm 3cm},clip,width=0.85\columnwidth]{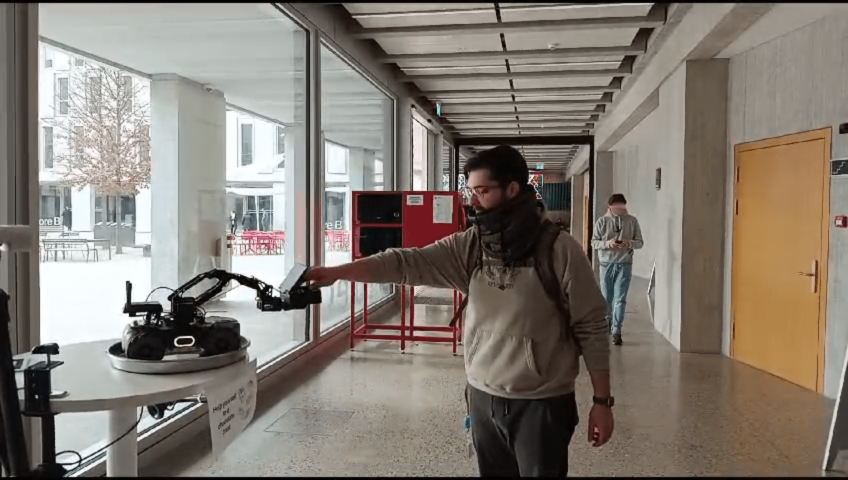}}%
\\[10 pt]\frame{\includegraphics[width=0.85\columnwidth]{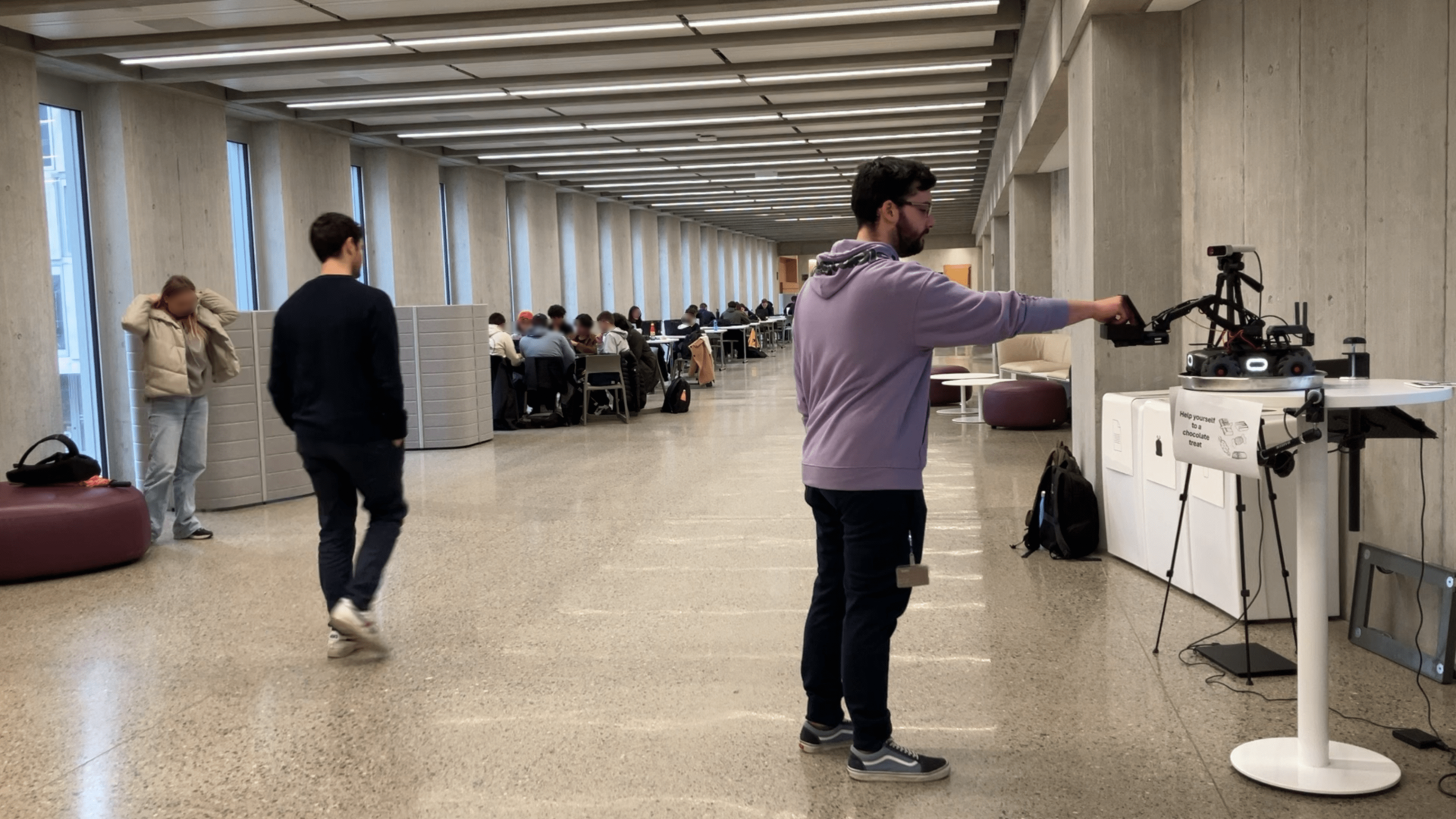}}%
\caption{Our robot is tested in two different public environments: it must track multiple nearby people, predict their intention to interact, and proactively offer them chocolate treats.}%
\label{fig:environments}
\end{figure}

To build socially compliant agents, it is crucial to appropriately detect and interpret the human intentions. 
In this way, robots can realize proactive behaviors and offer the requested service with high user satisfaction.  
Consider the concrete example of a robot tasked to distribute leaflets in a 
street, or offer chocolate treats to the visitors of a public building, see Fig.~\ref{fig:environments}.
The service robot should have the capability of ($i$) detecting nearby users, ($ii$) predicting their intentions, and ($iii$) promptly reacting to interested people.
In this way, people can feel involved and also the skeptical, suspicious, or shy user can resolve their doubt for a satisfactory \ac{hri} experience.
As a result, the service for which the robot is deployed can be efficiently delivered to people.

The detection of users entering the social space of the robot can be easily done with state-of-the-art sensors commonly used in robotics. 
Building on such solutions, in previous works~\cite{Abbate:ras:2024,Arreghini:icra:2024}, we designed a perception pipeline to discern the users' intention to interact. 
These self-supervised learning approaches use users' gaze cues and body motion information to predict the probability of future interaction.
Body posture can be used to classify the intention of multiple users with high accuracy and an advance time of more than $3$~s before the actual interaction~\cite{Abbate:ras:2024}.
The integration of facial features and a mutual gaze detector (developed on purpose for HRI applications~\cite{Arreghini:hri:2024}) further improves the perception performance~\cite{Arreghini:icra:2024}.
Once the user intention is predicted, it can trigger appropriate robot behaviors.
This pipeline has been successfully tested in controlled environments or with actors.

The present work aims to test our algorithms in the wild, tackling the challenge of dealing with real social environments with people unaware of the robot's presence.
In doing so, we provide the community with a comprehensive analysis of our experiences, consisting of real \acp{hri} occurring at different locations of a university campus.
To this end, we designed the robotic service task of offering chocolate treats to passing people.
In particular, the work aims to ($i$) realize a robust framework at the standards of real-life scenarios and ($ii$) understand which robot behavior is most suited in such an application.
To achieve these objectives, we propose the following contributions.
The first is very practical and regards the realization of a robust \ac{hri} pipeline allowing long-lasting experiments in the wild.
Our framework is built on our previous work~\cite{Abbate:ras:2024,Arreghini:hri:2024,Arreghini:icra:2024}, which had to be fine-tuned to comply with the specific challenges of real-life scenarios.
Secondly, we provide a rich dataset of people moving in public spaces and, possibly, interacting with the robot on which we perform detailed analyses of the users' behavior, the robotic perception, and the effects that these factors have on the interaction.
More in detail, we compare three different robot behaviors and analyze the intention of more than $1500$ people, in different locations. 
Finally, we derive conclusions from our real \ac{hri} experiences and draw lines for the future of robotic service applications, discussing the unsolved challenges awaiting researchers in social \ac{hri}.
The remainder of the paper is organized as follows. 
Sec.~\ref{sec:related_work} describes the state-of-the-art, Sec.~\ref{sec:experimental_setup} shows the setup used for putting our robot into the wild, while Sec.~\ref{sec:dataset} explains the dataset. 
The results of our real-life experiences are outlined in Sec.~\ref{sec:results}; final discussions and conclusions are in Sec.~\ref{sec:conclusions}.
%
\section{Related work}\label{sec:related_work}
In the context of \ac{hri}, social robots are expected to interact as peers with human companions~\cite{goodrich2008human}.
Leaving robots to operate autonomously in the wild poses multiple challenges that vary with the chosen application.
Indeed, the adoption of such systems is normally focused only on specific use cases, and not on the general consumer market~\cite{henschel2021makes}.
Furthermore, the majority of research in \ac{hri} has been conducted in controlled conditions where usually a single person interacts with a single robotic agent, see, e.g.~\cite{Baraglia:hri:2016}. 
However, for the test in real social scenarios, it is important to consider crucial aspects, such as the influence that robots have on their social environment, even beyond the person with whom it is directly interacting~\cite{jung2018robots} or the consequence that novelty effect can play on people's behavior around them~\cite{reimann2023social}.
Creating well-perceived and ever-improving \ac{hri} applications is an iterative process between designing and deployment that must keep a ``human first'' approach~\cite{willis2018human}. 
The process should start with the environment and problem selection, followed by understanding the users' behavior and needs and finally adjusting the reaction to provide socially acceptable \ac{hri}s.

In the context of real social scenarios, perceiving human needs is tightly related to understanding human intentions, a field where non-verbal communication cues are of utmost importance~\cite{Gasteiger:ijsr:2021}.
This topic has been widely studied for different \ac{hri} contexts: navigation~\cite{Agand:icra:2022}, collaborative tasks~\cite{Belardinelli:iros:2022,Vinanzi:icdler:2019} and social behavior interpretation~\cite{Zaraki:icrm:2014,Gaschler:iros:2012}.
Particularly useful for promptly reacting in the early stages of a \ac{hri} is detecting users' intention to interact, e.g. using only body posture information~\cite{Abbate:ras:2024}.
Human intention detection can be improved by adding gaze cues, widely considered a powerful indicator of a user's intentions~\cite{belardinelli_gaze-based_2023}.
This fusion of these two information sources resulted in our previous work~\cite{Arreghini:icra:2024}, which we also aimed to deploy ``in the wild''. 
Nonetheless, non-verbal communication between humans and robots can also flow in the opposite direction with the robot communicating its intentions to humans~\cite{Saunderson:ijsr:2019}. 
%
\diff{Most of the \ac{hri} studies are carried out in controlled environments, e.g., classrooms~\cite{woo2021use} or laboratories~\cite{ramis2020using}. 
Although invaluable, these studies fall short of addressing the complexities of the real-world~\cite{jung2018robots}, as they do not deal with challenges 
of dynamic environments, 
different users, users' reactions, and 
privacy issues~\cite{lutz2019privacy}. 
This last point can make robots deployment in 
public spaces very challenging, yet critical for social robotics advancement.} 

\section{Experimental setup}\label{sec:experimental_setup}

To collect meaningful \ac{hri} data in the wild, we design the task of offering chocolate treats to people walking in the surroundings of a robot.
A system of this kind, autonomously operating among people in real social contexts, presents technological challenges. 
The general experimental setup must be robust, reliable, and safe enough to deal with people not accustomed to interacting with robots.   
Both hardware and the robot behaviors needed to achieve such real-world experiences are described in what follows.
Also, we provide instructions to make our experimental setup reproducible.

\subsection{Hardware setup}\label{sec:hardware_setup}
\begin{figure}[!t]
    \vspace{2mm}
    \centering
    \begin{minipage}{0.46\columnwidth}    
        \frame{\includegraphics[trim={0cm 0.2cm 0.5cm 0.5cm},clip,width=\linewidth]{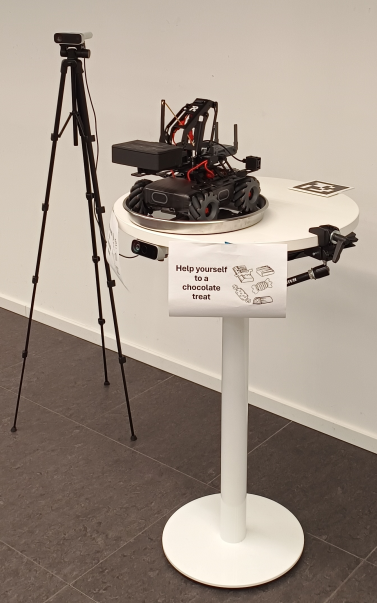}}
        \\[7pt]
        \frame{\includegraphics[trim={0cm 6cm 0cm 5cm},clip,width=\linewidth]{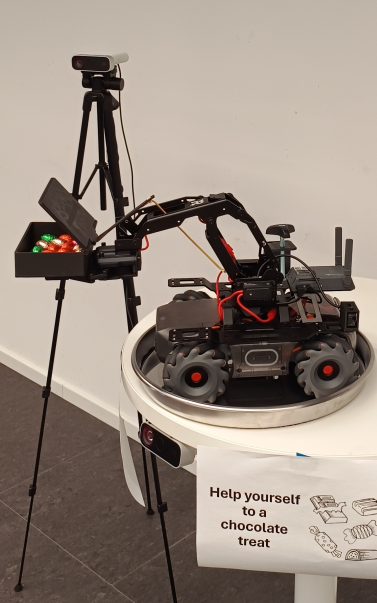}}
    \end{minipage}
    \hspace{1pt}
    \begin{minipage}{0.49\columnwidth} 
        \frame{\includegraphics[trim={5cm 5.5cm 0cm 0.0cm},clip,width=\linewidth]{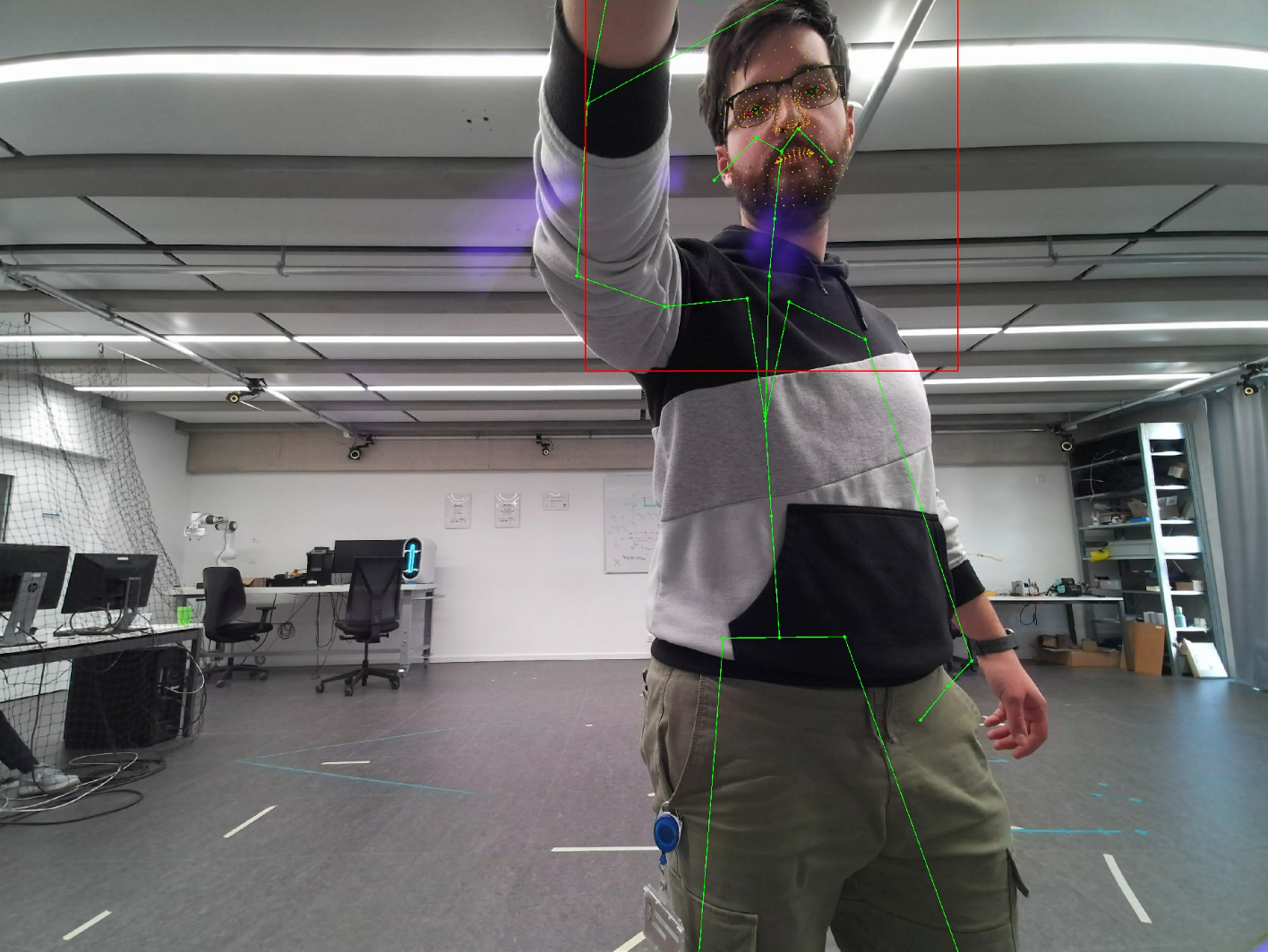}}
        \\[7pt]
        \frame{\includegraphics[trim={0cm 0.5cm 0cm 0.5cm},clip,width=\linewidth]{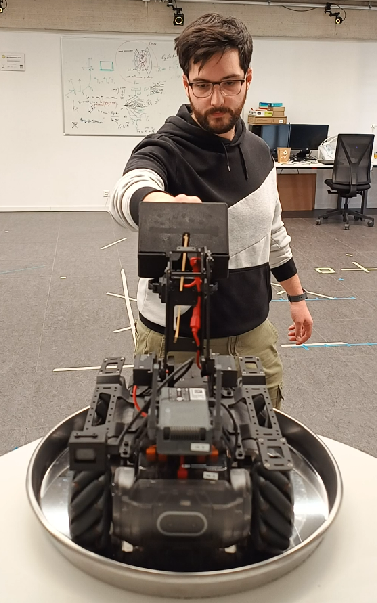}}
    \end{minipage}    
    \caption{Experimental setup (top left) with a zoomed view of the robot with the stretched arm offering chocolate treats (bottom left); a view from behind the robot interacting with a user (bottom right), and the same situation captured by the Robot Sensor (top right).}
    \label{fig:hardware_setup}
\end{figure}
The setup used in our work is shown in Fig.~\ref{fig:hardware_setup}.
We use a DJI RoboMaster EP Core\footnote{https://www.dji.com/ch/robomaster-ep-core}, a small-size mobile robot with omnidirectional wheels, LED lights, and a $2$ degrees of freedom arm. 
The arm is equipped with a simple gripper and has a payload capacity of \SI{0.3}{\kilogram}; we use it to hold a box with a lid that automatically opens when the arm stretches out. 
The robot is placed on a table \SI{1.10}{\metre} high; a border around the robot prevents it from accidentally falling.
Interaction with people is realized through gestural and visual communication, as the robot can: rotate in place toward nearby people; hand out things contained in the box by stretching its arm; display its active state through its LEDs.
We also use two Azure Kinect\footnote{https://learn.microsoft.com/en-us/azure/kinect-dk/system-requirements} RGB-D sensors to track people in the environment. 
They can stream $4$K images and provide built-in human body tracking.
The first one acts as the onboard robot sensor; thus, we call it \emph{Robot Sensor} and place it 
below the \diff{table} 
to match the robot's point of view. 
The second one, called \emph{Environment Sensor}, is 
used to track additional user data for 
\diff{subsequent analyses.} 
Its placement aims to cover the blind spots left by the Robot Sensor, 
mitigating the risk of user tracking loss \diff{if} 
individuals \diff{come} too close to the Robot Sensor.
Two dedicated laptops are connected to the sensors, running their driver and recording data on disk.
The one connected to the Robot Sensor also runs the code controlling the robot.
To implement different robot behaviors, described in Sec.~\ref{sec:robot_behaviors}, we use \diff{data} 
coming from the Robot Sensor. In particular, using body tracking data we compute the Euclidean distance of the users' torso; \diff{with RGB images, we obtain facial landmarks and gaze cues}. 
%
\diff{This information is fed to our interaction intention classifier, described in~\cite{Arreghini:icra:2024}, which continuously predicts if 
a person \diff{will interact}. 
This model uses 
a \ac{lstm} architecture to capture the inherent dynamic nature of people's decision process.}
Such a perception module has been adjusted to account for the detector jitter and noise, e.g., using a hysteresis thresholding. In this way, we prevent the robot from triggering repeatedly when the detector is used to control its behavior. 
\subsection{Robot behaviors}\label{sec:robot_behaviors}
We design and compare different robot behaviors to investigate their impact on the interaction with people in real-world scenarios.
At rest, the robot body is aligned with the Robot Sensor's forward direction, its LEDs are turned off, and the arm is retracted.
We define an \emph{offering motion} to a person as follows: the robot rotates toward the person, lights up its LEDs with the color yellow, and extends its arm. 
Automatically, the lid of the box held by the robot opens, allowing the person to take the content, in our case the chocolate treats. 
The robot keeps orienting itself towards the target user until the end of the interaction, which depends on the current robot's behavior (explained below).
When multiple people trigger the offering motion, the robot reacts to the closest one.
We define three robot behaviors:
\begin{description}[font=\normalfont, 
]
    \item[\emph{Passive}:] the offering motion is never activated and the robot stays still, with the arm stretched out and the box opened, allowing people to take a chocolate treat.
    \item[\emph{Distance-based} (in short, Distance):] the offering motion is triggered when the person's torso from the Robot Sensor is lower than \SI{1.5}{\metre}, regardless of the person's body orientation. The robot keeps following the target until the person moves further than the distance threshold.
    \item[\emph{Intention to Interact Detection-based} (IID):] the robot offering motion is triggered toward a person when their probability of interaction exceeds a threshold of $85\%$. The motion stops when this probability goes below a hysteresis threshold of $75\%$.
\end{description}
It is worth mentioning that the actual robot behavior does not aim to maximize the number of treats taken by people but rather to strive for \diff{a friendly and non-obtrusive interaction}.

\subsection{Data acquisition procedure}\label{sec:data_acquisition_procedure}
%
To collect real-life data, we select locations where we expect many people to pass by. 
We place the setup described in~\ref{sec:hardware_setup} in a way that does not obstruct the passage or force people to change their path (e.g., close to a wall in a large corridor).
The goal is to allow passers-by to avoid the robot if they do not plan to interact with it, without influencing their trajectory.
Then, we get a rough measure of the relative transformation between the Robot Sensor and the Environment Sensor. 
Such transformation is crucial to match people's trajectories acquired by the two sensors.
The two laptops and their wires are arranged to reduce visual clutter as much as possible.  
We fill the box held by the robot with chocolate treats, start the acquisition, and leave the system unattended. 
We stay in an area that is close enough to observe the experiment and allows us to remain unnoticed by passing people (e.g., sitting at a table in the vicinity of the setup). 
We want to limit our interference with passers-by as much as possible. 
To inform people that chocolate treats are being offered, two notice signs citing ``Help yourself to a chocolate treat'' are attached at both sides of the table (see Fig.~\ref{fig:hardware_setup}).
The program controlling the robot cycles between the three behaviors, autonomously switching every \SI{5}{\minute}.
Thus, no external input is needed. 
We only check the chocolate box for refill every \SI{15}{\minute}
after covering the two sensors to avoid recording undesired data during this operation.

\begin{figure}[!t]
\includegraphics[width=\columnwidth]{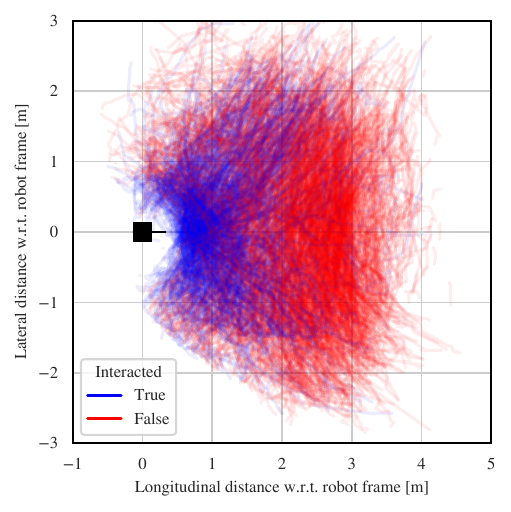}
    \caption{Trajectories of users who picked a treat (blue) or not (red), as tracked by the Environment Sensor. \diff{Blue trajectories concentrate around the robot, whereas red trajectories are uniformly distributed in space.}}    \label{fig:dataset_top_view_boolean}
\end{figure}
%
\section{Real-life dataset}\label{sec:dataset}

The data acquisition campaign is carried out following relevant ethical guidelines and approved by the local ethics committee of the University of Applied Sciences and Arts of Southern Switzerland (SUPSI).
We acquire data for three days in two different environments. 
Due to privacy concerns, we record only non-identifying data, such as people's torso trajectories and the internal state of our system. 
We collect a total of $5$~hours and $7$~minutes of data, with $1777$ people tracked in the robot's proximity.
Notably, no subjective evaluation measures from the users' are available due to two main factors.
First, asking every interacting user to fill out a survey poses a huge logistic challenge.
Second, the researchers' presence close to the robot undoubtedly has repercussions on the naturalness of the users' behavior towards the robot.
This important last point is further discussed in Sec.~\ref{sec:considerations}. 

We collect data in two different locations at the premises of the University Campus Est in Lugano, Switzerland.
In the first scenario (Fig.~\ref{fig:environments}, top), 
\diff{the robot is on the ground floor, at} the entrance of the campus canteen 
that is frequented by students, faculty staff, and occasional \diff{visitors}. 
Thus, the demographics for this sample include individuals 
\diff{from 19-year-old students to older adults}.
The second scenario (Fig.~\ref{fig:environments}, bottom) 
has a limited pool of people's ages, as 
\diff{the robot is in a corridor between classrooms, mainly attended by students.}

We collect data \diff{from both the Robot and Environment Sensor.} 
Their built-in body tracking system gives \diff{each tracked user a unique ID}. 
%
%
%
In particular, we collect two kinds of information.
The first is related to the Robot Sensor and consists of timestamps, user ID, 3D pose of the user's torso in the Robot Sensor frame, and \diff{the interaction intention detector output}. 
The latter contains timestamps, user ID, and 3D poses of the user's torso and hands \diff{in the Environment Sensor frame.} 
The 2D torso positions of people crossing the Environment Sensor field of view are shown in Fig.~\ref{fig:dataset_top_view_boolean}.
%
\diff{Besides} users' information, we 
record data about the robot's state and actions, \diff{e.g.} 
the currently selected robot behavior (explained in Sec.~\ref{sec:robot_behaviors}) and its state, i.e., if it is idle or performing an offering motion.  
From this data, we extract \diff{relevant information:} 
\emph{Pick Motion}, i.e., the first time each user moves a hand within \SI{0.3}{\metre} from the box position; \emph{Robot Offer}, i.e., any time the robot starts performing an offering motion; 
\emph{Successful Offer}, when a Pick Motion occurs within \SI{6}{\second} from a Robot Offer start.
Note that if the robot switches the target person during an offering motion, we count another Robot Offer. 
Also, we count Robot Offers and Successful Offers only when the robot is not deploying the Passive behavior.
The dataset is publicly available at \url{https://zenodo.org/records/12773705}.

\section{Results}\label{sec:results}

\diff{
The experimental setup and some interactions occurred during the experiments are presented in the accompanying video available at \url{https://youtu.be/NNgbNRxm5V4}.}

\subsection{Overall statistics}\label{sec:stats}

\begin{table}[!t]
    \setlength{\tabcolsep}{1.5pt}
    \centering
    \caption{Dataset Statistics}
    \label{tab:metrics}
    \begin{tabular}{lcccc}  
        \toprule
        & \multicolumn{3}{c}{\textbf{Robot Behavior}} & \\
        \cmidrule(lr){2-4}
        \textbf{Metric} & \textbf{Passive} & \textbf{Distance} & \textbf{IID} & \textbf{Total} \\
        \midrule
        Number of \textbf{tracked people} & 351 & 695 & 777 & 1777$^\ast$ \\ [2pt]
        Number of \textbf{Pick Motions} & 54 & 134 & 143 & 331 \\
        Pick Motions over tracked people & 15.4\% & 19.3\% & 18.4\% & {-} \\ [2pt]
        Number of \textbf{Robot Offers} & {-} & 184 & 254 & 438 \\
        Robot Offers over tracked people & {-} & 26.5\% & 32.7\% & {-} \\ [2pt]
        Number of \textbf{Successful Offers} & {-} & 66 & 114 & 180 \\
        Successful Offers over Robot Offers & {-} & 35.9\% & 44.9\% & 41.1\% \\
        Successful Offers over tracked people & {-} & 9.5\% & 14.7\% & {-} \\
        \midrule
        \multicolumn{5}{l}{{\scriptsize $^\ast$The people tracked in the single behaviors might not sum up to Total as some}}
        \\[-2pt]
        \multicolumn{5}{l}{{\scriptsize users might fall into two behaviors if tracked during a change of behavior.}} \\
        \bottomrule
    \end{tabular}
\end{table}
Table~\ref{tab:metrics} summarizes the dataset statistics according to the metrics defined in Sec.~\ref{sec:dataset}, providing interesting insights into the collected data.
Firstly, active robot behaviors (i.e., Distance and IID) yield, on average, more Pick Motions than the Passive one. 
Secondly, during the IID behavior, we can observe more Robot Offers over the tracked people than during the Distance behavior ($32.69\%$ and $26.47\%$, respectively).
Similarly, the value of Robot Offers made with IID also displays a higher rate of Successful Offers over tracked people ($14.67\%$) compared to the Distance behavior ($9.50\%$).
This comparison indicates that the offers triggered based on the users' intention are generally better targeted and thus better received by people, increasing the acceptance rate.

\begin{figure}    
    \includegraphics[width=\linewidth]{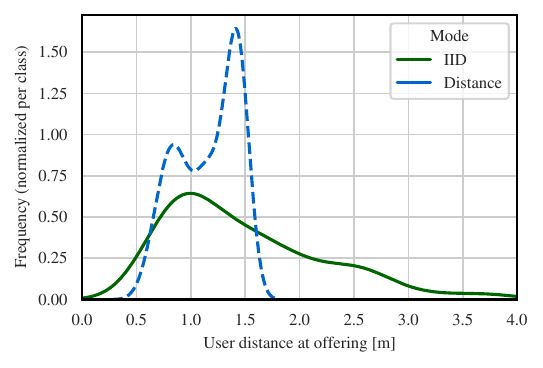}
    \caption{Distribution of the users' torso distance from the robot at the instant of Robot Offer. \diff{The IID case shows a broader distribution of users' distance at offering compared to the Distance case.}}
    \label{fig:trajectories_after_offering}
\end{figure}

The analysis on the user distance at which an offering is initiated shows an average for the IID behavior of \SI{1.52}{\metre}, consolidating our choice to set the threshold at \SI{1.5}{\metre} in the Distance behavior.
A more in-depth analysis of the distribution of the users' distances at the Robot Offer is reported in Fig.~\ref{fig:trajectories_after_offering}: the IID behavior (green) can initiate the offer when the user's position is much farther than \SI{1.5}{\metre}. 

\begin{figure}    
    \includegraphics[width=\linewidth]{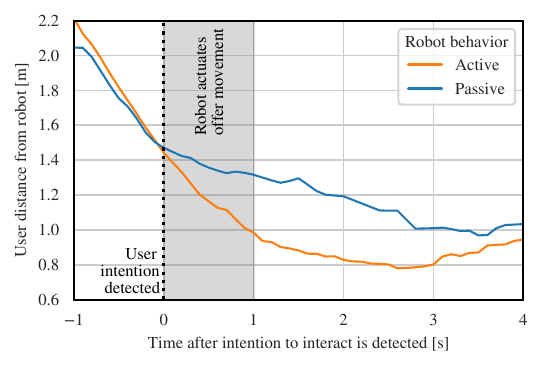}
    \caption{Median user distance to robot vs time.  The intention to interact is detected at time $t=0$ (vertical dotted line). For people interacting with the robot in any active behavior,  at $t=0$ the robot starts the offering motion, carried out in a time interval approximately corresponding to the shaded area. For users of the Passive robot (which does not move), $t=0$ is the time when the robot would have triggered if it had been Active.}
    \label{fig:distance_over_time}
\end{figure}

Similarly, differences between active (Distance or IID) and Passive behavior 
can be seen \diff{from approaching people motions} (Fig.~\ref{fig:distance_over_time}).
%
\diff{We} compare users towards whom the robot actuated the offering motion (orange), with a control group of users towards which the robot did not move because it was \diff{passive} 
(blue)--but would otherwise have actuated the offer. 
This comparison between such users \diff{aims} 
at isolating the impact of the robot's reaction on people's behavior. 
All trajectories are synchronized such that the detection occurs at time $t=0$. 
As expected, the user behavior in the two cases is similar until the instant of the robot's reaction. 
Then, users who experience the robot's offer tend to move closer and quicker toward the robot compared to the control group, who instead \diff{show} 
a more skeptical behavior with people \diff{hesitant to come close} and stay in the robot's proximity for long.

Further considerations about people's behavior can be done by looking at their torso velocity distributions in the case of interacting or non-interacting users.
\begin{figure}    
    \includegraphics[width=\linewidth]{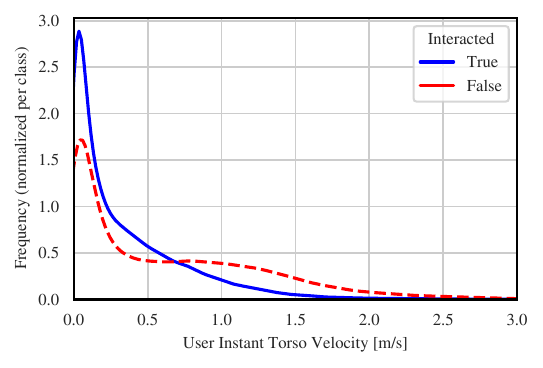}
    \caption{Users torso velocity distribution measured by Environment Sensor.}
    \label{fig:velocity_distribution}
\end{figure}
This analysis is reported in Fig.~\ref{fig:velocity_distribution} and shows that interacting people display a velocity distribution with a single peak very close to zero. This is likely due to the long time spent stationary in front of the robot during the Pick Motion phase of the interaction. 
The same distribution for non-interacting subjects shows a similar peak around zero, but a second shorter yet wider peak around $0.8$ m/s, which could be considered a slow walking speed. 
This could be explained by the non-interacting people displaying two different reactions towards the robot: indifference (passing unfazed by the robot's presence) or curiosity (stopping at the setup but ultimately avoiding interaction).

\subsection{Consideration and discussion}\label{sec:considerations}
Interesting anecdotal evaluations of the people's behavior observed during the acquisition campaign are listed below.
\subsubsection{Experience with robot} a previous experience with robots, or lack thereof, plays a substantial role in people's attitude towards the robot and willingness to interact with it.
Colleagues with known previous experience show to be more confident compared to people without previous experience.

\subsubsection{Reaction expectation} people who see the robot moving once, afterward expect the robot to always react to their presence, finding the Passive behavior underwhelming, in some cases thinking that the robot is malfunctioning.

\subsubsection{Age} users' age has a great influence on the potential users' attitudes towards the robot. 
Indeed, older adults and elderly people showed, in a few cases, initial skepticism toward the robot and seemed very unsure about its behavior.

\subsubsection{Researcher presence} the presence of researchers in proximity to the robot alters people's behavior who avoid coming closer to the robot. 

\begin{figure*}[t]   
    \centering
    \includegraphics[width=0.96\textwidth]{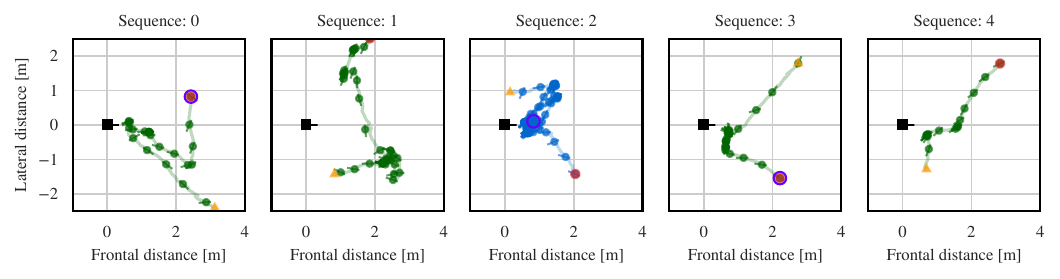}
    \caption{Examples of user trajectories. The green ones are performed during the IID behavior, the blue ones occurred during the Distance behavior. The red circle is the start of the trajectory, the yellow triangle the end, and the purple circle denotes the user position when the offering motion is performed.}\label{fig:interesting_trajectories_trails}
\end{figure*}
%
Figure~\ref{fig:interesting_trajectories_trails} shows example behaviors emerging from a sample of users' trajectories.
In sequence $0$, a user is initially crossing the robot's area, but then stops to come back and eventually interacts with the robot.
Sequence $1$ shows a similar initial behavior of the user, who simply walks in front of the robot. 
However, in this case, even though the user stops to check the setup, they do not interact with the robot.
The user in sequence $2$ remains in front of the robot for an extended amount of time slightly moving around; this trajectory could belong to a user who, attracted by the robot's behavior, toyed with it to challenge its tracking skill.
Sequence $3$ shows a case of seamless interaction; sequence $4$ corresponds to the motion of a user completely uninterested.

Finally, it is worth mentioning that, in general, the task and the robot, carefully designed to be as friendly as possible, triggered positive comments from the people passing by, indicating users' approval of the whole setup.
As mentioned, due to ethical and privacy issues, no extensive user data (that might include personal information of the users) could be collected. 
Gathering more complete data from unaware users would have been a serious privacy breach and singularly asking every person passing in front of the robot for consent or a subjective evaluation is logistically unfeasible.
%

\section{Conclusions and future work}\label{sec:conclusions}

We have created a robust setup capable of meaningful interactions with people in the wild for an extended amount of time.
During operation, the system \diff{can gather} 
data about the people passing in its proximity and more in-depth about the ones with whom it interacts.
%
\diff{The dataset is publicly available and}
contains information from $1777$ people tracked in more than \diff{\SI{5}{\hour}} of operation across $3$ days and $2$ environments. 
%
\diff{Our} analysis shows that a targeted reaction from a robot can 
\diff{greatly affect people's attitudes and behavior.}
%
\diff{This} aspect is reflected in the different success rates for the offers done in the different behaviors 
\diff{and} in the motion patterns of people passing in the robot surroundings.
However, our analyses \diff{use} 
simple objective user metrics. 
Indeed, privacy concerns for unaware people constitute a big issue in collecting more detailed personal data.
The solution of stopping passing people to ask for consent or a subjective evaluation questionnaire is unfeasible due to the experiments' ``in the wild'' nature and the large number of people involved.
\diff{Thus, future efforts should aim at} 
designing better evaluation procedures and metrics, \diff{for} 
a more comprehensive \diff{interactions analysis} 
while preserving privacy. This is instrumental in further improving perception and interaction skills of social robots.
%

\bibliographystyle{IEEEtran}
\bibliography{bibliography.bib}

\begin{thebibliography}{10}
\providecommand{\url}[1]{#1}
\csname url@samestyle\endcsname
\providecommand{\newblock}{\relax}
\providecommand{\bibinfo}[2]{#2}
\providecommand{\BIBentrySTDinterwordspacing}{\spaceskip=0pt\relax}
\providecommand{\BIBentryALTinterwordstretchfactor}{4}
\providecommand{\BIBentryALTinterwordspacing}{\spaceskip=\fontdimen2\font plus
\BIBentryALTinterwordstretchfactor\fontdimen3\font minus
  \fontdimen4\font\relax}
\providecommand{\BIBforeignlanguage}[2]{{%
\expandafter\ifx\csname l@#1\endcsname\relax
\typeout{** WARNING: IEEEtran.bst: No hyphenation pattern has been}%
\typeout{** loaded for the language `#1'. Using the pattern for}%
\typeout{** the default language instead.}%
\else
\language=\csname l@#1\endcsname
\fi
#2}}
\providecommand{\BIBdecl}{\relax}
\BIBdecl

\bibitem{Benitti:cae:2012}
F.~B.~V. Benitti, ``Exploring the educational potential of robotics in schools:
  A systematic review,'' \emph{Comp. \& Education}, vol.~58, no.~3, pp.
  978--988, 2012.

\bibitem{Gonzalez:as:2021}
C.~S. Gonz{\'a}lez-Gonz{\'a}lez, V.~Violant-Holz, and R.~M. Gil-Iranzo,
  ``Social robots in hospitals: a systematic review,'' \emph{Appl. Sci.},
  vol.~11, no.~13, p. 5976, 2021.

\bibitem{Choi:jhmm:2020}
M.~M.~O. Youngjoon~Choi, Miju~Choi and S.~S. Kim, ``Service robots in hotels:
  understanding the service quality perceptions of human-robot interaction,''
  \emph{J. of Hospitality Marketing \& Management}, vol.~29, no.~6, pp.
  613--635, 2020.

\bibitem{sirithunge2019proactive}
C.~Sirithunge, A.~B.~P. Jayasekara, and D.~Chandima, ``Proactive robots with
  the perception of nonverbal human behavior: A review,'' \emph{IEEE Access},
  vol.~7, pp. 77\,308--77\,327, 2019.

\bibitem{Zaraki:icrm:2014}
A.~Zaraki, M.~Giuliani, M.~B. Dehkordi, D.~Mazzei, A.~D'ursi, and D.~De~Rossi,
  ``An {RGB-D} based social behavior interpretation system for a humanoid
  social robot,'' in \emph{RSI/ISM Int. Conf. on Robot. and Mechatronics},
  2014, pp. 185--190.

\bibitem{nocentini2019survey}
O.~Nocentini, L.~Fiorini, G.~Acerbi, A.~Sorrentino, G.~Mancioppi, and
  F.~Cavallo, ``A survey of behavioral models for social robots,''
  \emph{Robotics}, vol.~8, no.~3, p.~54, 2019.

\bibitem{Abbate:ras:2024}
G.~Abbate, A.~Giusti, V.~Schmuck, O.~Celiktutan, and A.~Paolillo,
  ``Self-supervised prediction of the intention to interact with a service
  robot,'' \emph{Robot. Auton. Syst.}, vol. 171, p. 104568, 2024.

\bibitem{Arreghini:icra:2024}
S.~Arreghini, G.~Abbate, A.~Giusti, and A.~Paolillo, ``Predicting the intention
  to interact with a service robot: the role of gaze cues,'' in \emph{IEEE Int.
  Conf. Robot. and Autom.}, 2024, pp.~--.

\bibitem{Arreghini:hri:2024}
{S. Arreghini, G. Abbate, A. Giusti, and A. Paolillo}, ``A long-range mutual
  gaze detector for {HRI},'' in \emph{ACM/IEEE Int. Conf. Human-Robot Int.},
  2024, pp. 870--874.

\bibitem{goodrich2008human}
M.~A. Goodrich, A.~C. Schultz \emph{et~al.}, ``Human--robot interaction: a
  survey,'' \emph{Found. and Trends{\textregistered} in Human-Comp. Inter.},
  vol.~1, no.~3, pp. 203--275, 2008.

\bibitem{henschel2021makes}
A.~Henschel, G.~Laban, and E.~S. Cross, ``What makes a robot social? a review
  of social robots from science fiction to a home or hospital near you,''
  \emph{Curr. Rob. Reports}, vol.~2, pp. 9--19, 2021.

\bibitem{Baraglia:hri:2016}
J.~Baraglia, M.~Cakmak, Y.~Nagai, R.~Rao, and M.~Asada, ``Initiative in robot
  assistance during collaborative task execution,'' in \emph{ACM/IEEE Int.
  Conf. Human-Robot Int.}, 2016, pp. 67--74.

\bibitem{jung2018robots}
M.~Jung and P.~Hinds, ``Robots in the wild: A time for more robust theories of
  human-robot interaction,'' pp. 1--5, 2018.

\bibitem{reimann2023social}
M.~Reimann, J.~van~de Graaf, N.~van Gulik, S.~Van De~Sanden, T.~Verhagen, and
  K.~Hindriks, ``Social robots in the wild and the novelty effect,'' in
  \emph{Int. Conf. on Social Robot.}, 2023, pp. 38--48.

\bibitem{willis2018human}
M.~Willis, ``Human centered robotics: designing valuable experiences for social
  robots,'' in \emph{Proceedings of HRI2018 Workshop (Social Robots in the
  Wild). ACM, New York}, 2018.

\bibitem{Gasteiger:ijsr:2021}
N.~Gasteiger, M.~Hellou, and H.~S. Ahn, ``Factors for personalization and
  localization to optimize human-robot interaction: A literature review,''
  \emph{Int. J. of Social Robot.}, pp. 1--13, 2021.

\bibitem{Agand:icra:2022}
P.~Agand, M.~Taherahmadi, A.~Lim, and M.~Chen, ``Human {Navigational} {Intent}
  {Inference} with {Probabilistic} and {Optimal} {Approaches},'' in \emph{IEEE
  Int. Conf. Robot. and Autom.}, 2022, pp. 8562--8568.

\bibitem{Belardinelli:iros:2022}
A.~Belardinelli, A.~R. Kondapally, D.~Ruiken, D.~Tanneberg, and T.~Watabe,
  ``Intention estimation from gaze and motion features for human-robot
  shared-control object manipulation,'' in \emph{IEEE/RSJ Int. Conf.
  Intelligent Robots Sys.}, 2022, pp. 9806--9813.

\bibitem{Vinanzi:icdler:2019}
S.~Vinanzi, C.~Goerick, and A.~Cangelosi, ``Mindreading for {Robots}:
  {Predicting} {Intentions} via {Dynamical} {Clustering} of {Human}
  {Postures},'' in \emph{Int. Conf. on Dev. and Learn. and Epigen. Rob.}, 2019,
  pp. 272--277.

\bibitem{Gaschler:iros:2012}
A.~Gaschler, S.~Jentzsch, M.~Giuliani, K.~Huth, J.~de~Ruiter, and A.~Knoll,
  ``Social behavior recognition using body posture and head pose for
  human-robot interaction,'' in \emph{IEEE/RSJ Int. Conf. Intelligent Robots
  Sys.}, 2012, pp. 2128--2133.

\bibitem{belardinelli_gaze-based_2023}
A.~Belardinelli, ``Gaze-based intention estimation: principles, methodologies,
  and applications in {HRI},'' \emph{ACM Trans. on Human-Robot Int.}, 2023.

\bibitem{Saunderson:ijsr:2019}
S.~Saunderson and G.~Nejat, ``How robots influence humans: A survey of
  nonverbal communication in social human-robot interaction,'' \emph{Int. J. of
  Social Robot.}, vol.~11, pp. 575--608, 2019.

\bibitem{woo2021use}
H.~Woo, G.~K. LeTendre, T.~Pham-Shouse, and Y.~Xiong, ``The use of social
  robots in classrooms: A review of field-based studies,'' \emph{Edu. Res.
  Rev.}, vol.~33, p. 100388, 2021.

\bibitem{ramis2020using}
S.~Ramis, J.~M. Buades, and F.~J. Perales, ``Using a social robot to evaluate
  facial expressions in the wild,'' \emph{Sensors}, vol.~20, no.~23, p. 6716,
  2020.

\bibitem{lutz2019privacy}
C.~Lutz, M.~Sch{\"o}ttler, and C.~P. Hoffmann, ``The privacy implications of
  social robots: Scoping review and expert interviews,'' \emph{Mobile Media \&
  Communication}, vol.~7, no.~3, pp. 412--434, 2019.

\end{thebibliography}

\end{document}